\documentclass[sigconf]{acmart}
\usepackage{subcaption}
\usepackage{placeins}
\usepackage{multicol}
\usepackage{dblfloatfix}

\AtBeginDocument{%
  \providecommand\BibTeX{{%
    \normalfont B\kern-0.5em{\scshape i\kern-0.25em b}\kern-0.8em\TeX}}}

\begin{document}

\title{Enhancing crowd flow prediction in various spatial and temporal granularities}

\author{Marco Cardia}
\email{marco.cardia@phd.unipi.it}
\orcid{0000-0002-0458-589X}
\affiliation{%
  \institution{University of Pisa}
  \streetaddress{Lungarno Antonio Pacinotti, 43}
  \city{Pisa}
  \country{Italy}
  \postcode{56126}
}

\author{Massimiliano Luca}
\email{mluca@fbk.eu}
\orcid{0000-0001-6964-9877}
\affiliation{%
  \institution{Free University of Bolzano}
  \streetaddress{Piazza Domenicani, 3}
  \city{Bolzano}
  \country{Italy}
  \postcode{39100}
}
\affiliation{%
  \institution{Bruno Kessler Foundation (FBK)}
  \streetaddress{Via Sommarive, 19}
  \city{Trento}
  \country{Italy}
  \postcode{38123}
}

\author{Luca Pappalardo}
\email{luca.pappalardo@isti.cnr.it}
\orcid{0000-0002-1547-6007}
\affiliation{%
  \institution{Institute of Information Science and Technology (ISTI), National Research Council (CNR)}
  \city{Pisa}
  \country{Italy}
}

\renewcommand{\shortauthors}{Cardia, et al.}

\begin{abstract}
Thanks to the diffusion of the Internet of Things, nowadays it is possible to sense human mobility almost in real time using unconventional methods (e.g., number of bikes in a bike station). Due to the diffusion of such technologies, the last years have witnessed a significant growth of human mobility studies, motivated by their importance in a wide range of applications, from traffic management to public security and computational epidemiology. 
A mobility task that is becoming prominent is crowd flow prediction, i.e., forecasting aggregated incoming and outgoing flows in the locations of a geographic region.
Although several deep learning approaches have been proposed to solve this problem, their usage is limited to specific types of spatial tessellations and cannot provide sufficient explanations of their predictions.
We propose CrowdNet, a solution to crowd flow prediction based on graph convolutional networks. 
Compared with state-of-the-art solutions, CrowdNet can be used with regions of irregular shapes and provide meaningful explanations of the predicted crowd flows.
We conduct experiments on public data varying the spatio-temporal granularity of crowd flows to show the superiority of our model with respect to existing methods, and we investigate CrowdNet's reliability to missing or noisy input data.
Our model is a step forward in the design of reliable deep learning models to predict and explain human displacements in urban environments.
\end{abstract}




\maketitle

\section{Introduction}
Web technologies and the Internet of things have a predominant role in modeling human mobility. For instance, a recent survey \cite{luca2021survey} highlighted how data from social networks are widely used in studies discussing human mobility at an individual level. On the other side, it is also possible to use sensors to frequently capture aggregated mobility information. As an example, it is possible to measure where the origin and destination trips of shared bikes or to collect GPS data in real-time.
The study of human mobility is relevant to a large variety of topics, including public safety, migration, on-demand services, pollution monitoring, diffusion of epidemics, and traffic optimisation \cite{Barbosa2018,BoyceForecastingUrbanTravel,KrebMappingNetworksTerrorist,Nagel_1995,Tizzoni_HumanMobilityEpidemics,WangRoadUsagePaterns,khaidem2020optimizing,luca2021leveraging,luca2022modeling}.
For this reason, and thanks to the recent deluge of digital data and the striking development of artificial intelligence, there has been a vast scientific production on various tasks involving human mobility data \cite{luca2021survey, Barbosa2018}. 
A notable example is crowd flow prediction, consisting in forecasting the aggregated incoming and outgoing flows of people that move across regions in a geographic area \cite{luca2021survey, Zhang2016}. 
The main challenge in solving this task lies in capturing the close and far spatial and temporal dependencies in the data at the same time.
To date, crowd flow prediction is tackled with two main approaches: statistical models based on time series, which generally cannot capture both the spatial and the temporal dependencies; and models based on deep learning, which outperform traditional statistical models thanks to their complex architecture \cite{luca2021survey}.

Existing solutions to crowd flow prediction are not very meaningful to policymakers: as predictions are produced in form of a single value per location representing the aggregated inflow or outflow, no information is provided about the origin and the destination of these flows. 
However, this networked information is crucial to understand the density of movements throughout the city and its evolution in time, manage public events and emergency situations, maintain an efficient public transport system, and forecast the direction in which a viral disease may spread out \cite{iot2010003}.

In this paper, we propose CrowdNet, a deep learning approach that solves crowd flow prediction and overtakes the aforementioned limitations.
With respect to state-of-the-art approaches (e.g., STResNet \cite{Zhang2016}), our approach brings several advantages: 
\begin{itemize}
\item it predicts crowd inflows and outflows in the administrative areas of a city, while existing approaches can work on a regular tessellation (grid) of the territory only;
\item it corroborates the prediction with useful information such as the origin and destination of a crowd flow;
\item it solves also the flow prediction consisting in forecasting the flows \emph{among} geographic regions;
\end{itemize}

CrowdNet employs 1) a convolution-based graph neural network to model nearby and distant spatial dependencies between regions in a city, and 2) convolutional neural networks to capture temporal dependencies.
In particular, a Time Block captures the temporal dependencies and a Spatial Block captures the spatial dependencies in the data. 
The output of the Spatial Block is provided to another Time Block.
This structure defines a module named ST-Conv-Block.


We evaluate CrowdNet on different datasets describing the movements of bikes in New York City, taxis in Beijing and bikes in Washington DC varying both the size of the tessellation (i.e. the dimensions of the regions) and the time interval (i.e. the time slots in which the crowd flows are grouped).
CrowdNet achieves results that are comparable to other state-of-the-art solutions while providing richer predictions that work for both regular and irregular tessellations.
We also provide the code to reproduce CrowdNet and our experiments on public datasets at \url{https://github.com/jonpappalord/crowd\_flow\_prediction}.
The remainder of this paper is organized as follows. 
In Section \ref{sec:problem_definition}, we provide the backgrounds and we define both the crowd flow prediction problem and flow prediction problem. In Section \ref{sec:CrowdNet}, we present the deep learning model used to solve the previously defined problem.
In Section \ref{sec:experiments}, we described the used datasets, the evaluation metrics and the experimental settings.
In Section \ref{sec:results}, we provide the obtained results.
In Section \ref{sec:related_works}, we provide a brief description of the literature related to crowd flow prediction.
In Section \ref{sec:discussion_conclusions}, we give a summary of the contributions of our work and possible future improvements.

\section{Related works}
\label{sec:related_works}

Statistical-based methods for crowd flow prediction represent flows through equation matrices and adopt independent variables to represent adjacent areas and historical data. 
Autoregressive Integrated Moving Average (ARIMA) \cite{ARIMA} uses a number of lagged observations of univariate time series to forecast new observations.
Vector Auto-Regressive (VAR) exploits multiple time series to capture the pairwise relationships among flows \cite{hansen_1995}.
Overall, autoregression approaches cannot capture neither complex temporal and spatial dependencies and they require feature engineering to transform raw data into appropriate internal representations for spatio-temporal dependency detection. In contrast, Deep Learning (DL) approaches can discover features from raw data automatically \cite{Hinton504}.

\paragraph{Deep Learning approaches.}
There are many DL algorithms that are specifically designed to solve the crowd flow prediction problem. Many of them are collected in a recent survey by Luca et al. \cite{luca2021survey}. Most of the solutions leverage convolutional neural networks (CNNs) and recurrent networks (RNNs) to capture spatio-temporal patterns and dependencies. Examples are \cite{jiang2021deepcrowd,dai2021attention,wang2020seqst,Zhang2016,Tian2020,Ren2020,liu2020dynamic,Li2019DenselyCC,du2020,Yuan2020}. Some other solutions also rely on attention mechanisms. Examples are \cite{jiang2021deepcrowd,dai2021attention,wang2020seqst,Tian2020}. In what follows, we introduce additional details of the models we will use as baselines in this study.
DeepST \cite{Zhang} captures temporal patterns using the assumption that time series always respect temporal closeness, period, and seasonal trend. A convolutioanl neural network (CNN) module captures spatial dependencies, and a fusion mechanism combines the outputs. 
Spatio-Temporal Residual Network (STResNet) \cite{Zhang2016} improves on DeepST adding residual learning, a parametric and matrix-based fusion mechanism, and the consideration of external factors.
Local-Dilated Region-Shifting Network (LDRSN) \cite{Tian2020} combines local and dilated convolutions to learn the nearby and distant spatial dependency, which makes it more resilient to overfitting than approaches based on CNNs.
Hybrid-Integrated DL Spatio Temporal network (HIDLST) \cite{Ren2020} exploits an long-short term memory network (LSTM) to capture dynamic temporal dependency in time series and Residual CNNs to capture spatial dependencies.
Attentive Traffic Flow Machine (ATFM)\cite{liu2020dynamic} captures spatial-temporal dependencies with two convolutional LSTMs (ConvLSTM) units and an attention mechanism able to infer the trend evolution exploiting dynamic spatial-temporal feature representation learning.
Li et al. \cite{Li2019DenselyCC} proposes a model that is made up of densely connected CNNs to extract spatial characteristics, an attention-based long short-term memory module to capture temporal components and a fully connected neural network to extract features from external factors.
Deep Spatio-Temporal Irregular Convolutional Residual LSTM (DST-ICRL) \cite{du2020} integrates multi-channel traffic representations, irregular convolution residual networks and LSTMs to provide crowd flows forecasting. 
Multi-View Residual Attention Network (MV-RANet) \cite{Yuan2020} captures spatial dependencies by a double-branch residual attention network: one branch for small-scale dependency, the other one serves as an attention model, extracting spatial dependencies at large scale. External features are represented as three graphs of functional areas.

\section{Problem Definition}
\label{sec:problem_definition}
In this section, we formalise the problem of crowd flow prediction and introduce the main concepts used in the paper.

\begin{definition}[Spatial Tessellation]
Let $R$ be a geographical area and $G$ a set of polygons. $G$ is called \textit{tessellation} if the following properties hold:
\begin{enumerate}
    \item $G$ contains a finite number of polygons (i.e., tiles) $l_i$, so that $G = \{l_i: i=1, ..., n\}$;
    \item Locations are not overlapped, that is $l_i \cap l_j, \forall i \neq j$;
    \item The union of all the locations entirely covers $R$, i.e. $\bigcup_{i=1}^{n}l_i = R$.
\end{enumerate}

Tessellations allow us to map data points into a finite number of tiles within the area, instead of having raw positions expressed in coordinates.
Tiles are represented by either regular geometric shapes such as squares, triangles, quadrilaterals or hexagons, or irregular ones such as census cells or administrative units. 
Properties (2) and (3) ensure that each point is assigned to only one tile.
\end{definition}

Mobility flows represent aggregated movements among geographic locations, and they are usually represented as an Origin-Destination matrix.

\begin{definition}[Origin-Destination matrix]
    An Origin-Destination matrix is a matrix $T \in \mathbb{N}^{nxm}$ where $n$ is the number of different origin regions and $m$ is the number of distinct destination regions. $T_{i,j}$ denotes the number of individuals moving from region $i$ to region $j$.
\end{definition}

The origin and destination regions often coincides ($n=m$).
In the Crowd Flow Prediction problem, flows are aggregated into crowd flows (either incoming or outgoing) and represented as a bi-dimensional matrix in which an element represents the crowd flow in a tile during a certain time interval. 

\begin{definition}[Crowd Flow]
\label{def:inoutflow}
Given a trajectory $T_u$ describing the movements of an individual $u$, the set of \textit{tiles} intersected by $T_u$ in a time interval $\Delta t$ is defined as:
    \begin{equation}\label{eq:trajIntersect}
        q^t_{T_u} = \{(p_k \rightarrow t) \in \Delta t \wedge (p_k \rightarrow (x, y)) \in (i, j) | (i, j)\}
    \end{equation}
where the pair $(i, j)$ indicates a cell on an $I \times J$ grid and $p_k$ is $u$'s current location, identified by coordinates $(x,y)$.

Let $Q$ be the set of locations covered by all the individual trajectories, and let $t-1$, $t$ and $t+1$ be three consecutive time spans:
    \begin{itemize}
        \item The incoming crowd flow to a location $(i,j)$ is the number of individuals that were not in $(i,j)$ at time $t - 1$ and are in $(i, j)$ at time $t$.
        
        \begin{equation} \label{eq:inFlow}
        in_t^{(i,j)} = \sum_{T\in Q} |\{t > 1 : (i,j) \notin q_{T}^{t-1} \wedge (i,j) \in q_{T}^{t}\}|
        \end{equation}
        
        \item The outgoing crowd flow from a location $(i, j)$ is the number of individuals that were in $(i,j)$ at time $t$ and are no longer in $(i,j)$ at time $t + 1$.
        
        \begin{equation} \label{eq:outFlow}
        out_t^{(i,j)} = \sum_{T\in Q} |\{t > 1 : (i,j) \in q_{T}^{t} \wedge (i,j) \notin q_{T}^{t+1}\}|
        \end{equation}
    \end{itemize}

\end{definition}

Given the aforementioned, we define the problem as follows:

\begin{definition}[Crowd Flow Prediction]
\label{def:crowd_flow_prediction_problem}
Given a spatial tessellation $R$ composed by $n$ tiles and the crowd flows for each cell for $t$ time intervals, crowd flow prediction consists in forecasting $X_{t+c}$, where $c \in \mathbb{N}$, given the historical crowd flows $\{X_{i} : i = 1, ..., t\}$.
\end{definition}

A variant of crowd flow prediction is flow prediction:

\begin{definition}[Flow Prediction Problem]
    Given a spatial region $R$ and a temporal Origin-Destination matrix $T \in \mathbb{N}^{txnxm}$ where $n$ is the number of different origin tiles, $m$ is the number of distinct destination tiles and $t$ is the number of time intervals, flow prediction consists in predicting the next Origin-Destination matrix, i.e. the OD matrix at time $t+1$, given the historical flows $\{T_{i} : i = 1, ..., t\}$.
\end{definition}

Our model, presented in the next section, can solve both crowd flow prediction and flow prediction.

\section{CrowdNet}
\label{sec:CrowdNet}

CrowdNet is a deep neural network whose input is a temporal origin-destination matrix that describes historical flows among different regions, allowing it to use tessellations of various shapes (e.g., irregular tessellations).
First, the model solves flow prediction, forecasting the flows among all pairs of regions in the tessellation.
Then, it solves crowd flow prediction summing all the flows in the predicted OD matrix that have as a destination (origin) $k$, so to obtain the crowd inflow (outflow) of region $k$.

Given a geographic area tasselled into $n$ regions, we represent a set of flows at time $t$ as tensors $F_t \in \mathbb{R}^{n\times n}$, where the first dimension represents the origin and the second dimension represents the destination of the flow. 
Hence, $F_t (i,j)$ contains the flow at time $t$ moving from region $i$ to region $j$. 
A flow equal to $0$ means that no people move from region $i$ to region $j$ at time step $t$.

We adapted CrowdNet from the work by Yu et al. \cite{YuTrafficForecasting} on traffic forecasting. 
In particular, we treat the problem as a weighted link prediction applied to temporal dynamic directed graphs, where each node represents a region and each weighted edge quantifies the flow between two regions. 
Formally, a weighted graph, at the $t$-th time step, is a triple $G=(V,E_t,f_t)$, where $V$ is a set of vertices, $E_t$ is a set of edges, i.e., a set of ordered pairs $(u,v)$ where $(u,v) \in V \times V,$ with $u \neq v$ and $f_t$ is a function, $f_t: E_t \rightarrow \mathbb{N}$ assigning a value representing the weight of the edge \cite{HARARY199779}.
The network provides as output the graph at the $t+1$ time interval, i.e., the triple $G=(V, E_{t+1}, f_{t+1})$.

\subsection{Architecture}
We can formalise CrowdNet as: 
\begin{equation}
    \mbox{CrowdNet}(X_t, A) \rightarrow Y
\end{equation}
where $X_t \in \mathbb{R}^{n \times n \times k}$ are the OD matrices with $n$ nodes from time $t$ to $t + k$. $A \in \mathbb{R}^{n \times n}$ is the adjacency matrix of the graph (representing the Origin-Destination flows), which is:
\[
A_{i,j} = \left\{
  \begin{array}{lr}
    1 & \text{if } i \text{ and } j \text{ are linked in at least one time interval}\\
    0 & \text{otherwise}.
  \end{array}
\right.
\]

$Y \in \mathbb{R}^{n \times n \times l}$ is the model's prediction, where $l$ is the number of time intervals predicted and $n$ is the number of regions. 
Therefore, CrowdNet's predictions are the adjacency matrices from time $t+k+1$ to $t+k+l$. In our experiments, we fix $l=1$.
$Y$ is then aggregated into a bi-dimensional matrix where each element represents the inflow and the outflow, solving the crowd flow prediction problem.
Formally, the second output of CrowdNet is $Y' \in \mathbb{R}^n \times 2$ where $n = q \times q$, $q$ is the number of tiles in the x axis and $q$ is the number of tiles in the y axis. 

\begin{figure*}[!]
    \centering
    \includegraphics[width=.83\linewidth]{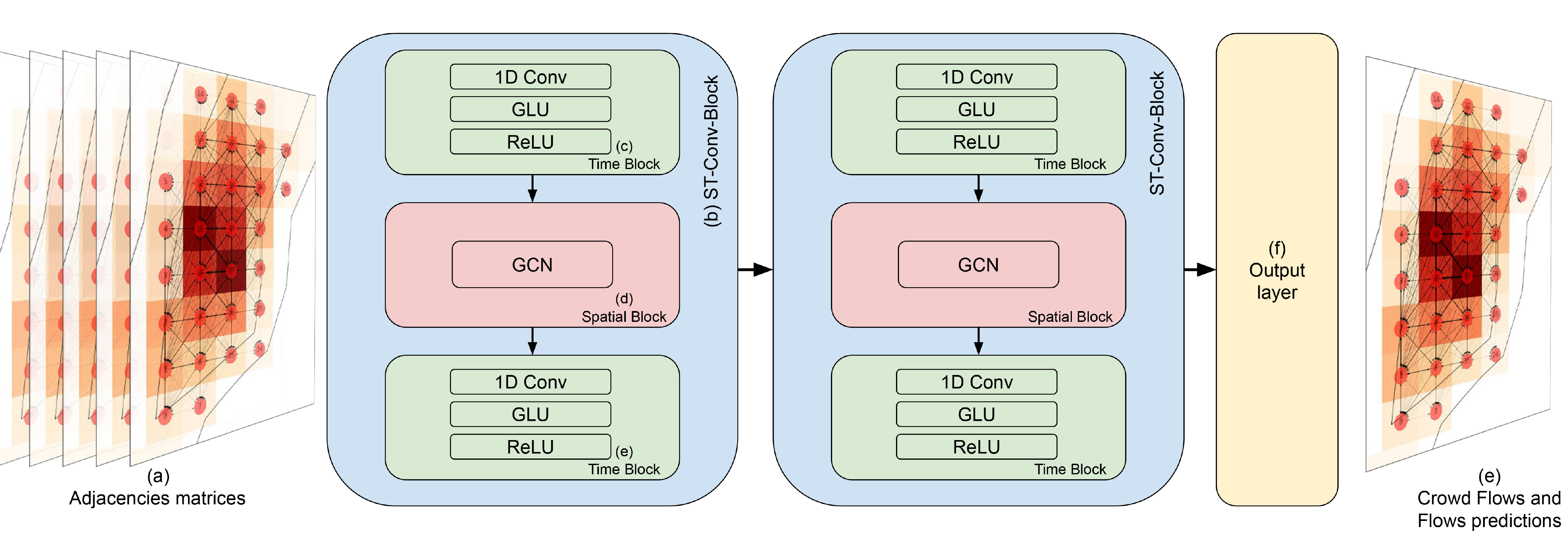}
    \caption{\small Architecture of CrowdNet. 
    }
    \label{fig:CrowdNetArch}
\end{figure*}

CrowdNet's architecture is composed of several spatio-temporal convolutional blocks, each made up of a multi-layer structure with two convolutional layers and one spatial graph convolutional layer in between. 
The former captures temporal dependencies and the latter catches the spatial dependencies.
Since for crowd flow prediction it is necessary a good response to dynamic changes \cite{YuTrafficForecasting}, we apply convolutions on the time axis to capture the temporal features of flows \cite{gehring2017convolutional}.
Figure \ref{fig:CrowdNetArch} schematizes CrowdNet's architecture.

\paragraph{Time Block.}

Inspired by Gehring et al. \cite{gehring2017convolutional}, we exploit CNNs to capture temporal dependencies.
This choice is justified by the fact that CNNs perform well when predicting flows dynamic changes. 
With respect to Recurrent Neural Networks (RNNs), CNNs have a faster training and the lack of dependencies constrains to previous steps allows a parallel and controllable training process through the multi-layer convolutional structure.

The Time Block (TB) contains a convolution followed by Gated Linear Units (GRUs) \cite{dauphin2017language}, which implement a gating mechanism over the output of the convolution. Such operations can be formalised as:
\begin{equation}
    \Gamma = (X * \Theta_1 +b_0) \circ \sigma (X * \Theta_2 + b_1)
\end{equation}
where $X$ is the input, $\Theta_1$, $\Theta_2$, $b_0$ and $b_1$ are learnable parameters, $\sigma$ is the sigmoid function, $*$ is the convolution operator, and $\circ$ denotes the Hadamard product.
The sigmoid gate controls what is relevant for discovering the structure and the dynamics of the time series.
To enable the use of deep convolutional networks, we use residual connections from the input $X$ to the output of each layer.
We use Rectified Linear Unit ($ReLU$) as the final activation function, defined as $ ReLU(x) = max(0,x)$. 
In summary:
\begin{equation}
    TB = ReLU(X * \Theta_3 + b_2 + \Gamma).
\end{equation}


\paragraph{Spatial Block.}
We define a Graph Convolution as:
\begin{equation}
    X' = \hat{D}^{-\frac{1}{2}} \hat{A}\hat{D}^{-\frac{1}{2}} X \Theta 
\end{equation}
where $ \hat{A} = A + I $, i.e., $\hat{A}$ is the adjacency matrix of the directed graph $G$ with self-loops. $I$ is the identity matrix. $\hat{D}$ is a diagonal matrix:
\begin{equation}
    \hat{D}_{i,i} = \sum_{j=0}^n \hat{A}_{i,j}
\end{equation}
where the element $(i,i)$ is the number of adjacent nodes for the node $i$. All the other elements are equal to $0$.
$X$ is the input, i.e., the origin-destination matrix for a defined time interval, as defined in Section \ref{sec:CrowdNet}.
As detailed in \cite{KipfGCN}, this operation is better motivated by a first-order approximation of localised spectral filters on the graph.

The operation of graph convolution is used by a layer block, named Spatial Block (SB). 
It is a one-layer Graph Convolutional Network (GCN) having the following form as forward model:
\begin{equation}
    SB = ReLU(\hat{D}^{-\frac{1}{2}} \hat{A}\hat{D}^{-\frac{1}{2}} X \Theta)
\end{equation} 
where $ReLU$ is the rectified linear unit activation function applied to the graph convolution operation. 


\paragraph{ST-GCN Block.}

The ST-GCN Block is composed of a Time Block, a Spatial Block, and another Time Block.
The Spatial Block is fed by the first Time Block and performs a graph convolution that can be expressed as:
\begin{equation}
    X'' = SB(X, \hat{A}) = ReLU(\hat{D}^{-\frac{1}{2}} \hat{A}\hat{D}^{-\frac{1}{2}} X \Theta)
\end{equation} 

The output of the Spatial Block layer is provided to a Time Block, which in turn returns: 
\begin{equation}
    X''' = TB(X'') = ReLU(X'' * \Theta_3 + b_2 + \Gamma).
\end{equation}

Finally, a batch normalisation is applied to the output of the the last Temporal Block.
Batch normalisation is defined as
\begin{equation}
     y' =  \frac{X''' - E[X''']}{\sqrt{Var(X''') + \epsilon}} * \gamma + \beta
\end{equation}
where $\gamma$ and $\beta$ are learnable parameters. It allows to use higher learning rates, to have faster training and permits to take less care to initialisation \cite{ioffe2015batch}.


To summarise, CrowdNet is composed of two ST-GCN layers as previously defined and one output layer.
The last block maps the outputs of the last ST-GCN layer into a single step prediction output.

The loss function used in CrowdNet is the Mean Squared Error (MSE), defined as:
\begin{equation}
    MSE = \frac{\sum_{i=1}^n(y_i-\hat{y_i})^2}{n}
\end{equation} 
where $y_i$ is the real value, $\hat{y_i}$ is the prediction, and $n$ is the size of the dataset.

\section{Experiments}
\label{sec:experiments}

In this section, we describe the dataset, the evaluation metrics, the baselines, and the experimental settings.

\subsection{Datasets}
\label{sec:datasets}
The \textit{Citi Bike System} dataset \footnote{https://ride.citibikenyc.com/system-data} describes trips recorded by the New York Official Bike sharing system from 2013 to date.
We consider trips from April to September 2014 because it is the range of dates usually used in the literature to test crowd flow prediction methods \cite{du2020, Li2019DenselyCC, liu2020dynamic, Tian2020, Zhang2016, Zhang}. Each record contains also the start and end times of the ride, and the coordinates of the starting and ending bike stations. 

The \textit{Taxi Beijing} dataset is based on T-Drive \cite{yuan2011driving}. It was collect by Microsoft in the area of Beijing, China and it contains the GPS location of 10,357 taxis sampled every 177 seconds. The data were collected over a period of one week in February 2008. 

The \textit{Capital Bikeshare} dataset \footnote{https://www.capitalbikeshare.com} describe the bike trips of the Washington D.C. bike sharing system. We consider the trips from January 2018 to January 2020. The information contained in each record are similar to the ones described for Bike in New York City and contains identifiers, latitude and longitude of the starting station and the ending station with the relative times.

\noindent \textbf{Preprocessing.} We use library scikit-mobility \cite{pappalardo2019scikitmobility} to construct a squared tessellation over New York City, Beijing and Washington D.C.. 
A squared tessellation is a division of a geographic area into equal-sized tiles.
Each tile is described by an identifier, the shape of the polygon describing the tile, and the position of the tile in a rectangular matrix modelling the squared tessellation. 



We use a spatial join to associate the stations' coordinates to the tile they fall within. Finally, we aggregate the joined dataset into an OD matrix and into a bi-dimensional matrix describing the crowd flows for each tile.
For example, given a time interval and a position $(i,j) \in \mathbb{R}^{n\times m}$ in the bi-dimensional map of size $n \times m$, the inflow of $(i,j)$ is the sum of all the flows having as destination the cell $(i,j)$.
Analogously, the outflow of $(i,j)$ is the sum of all the flows having as origin the cell $(i,j)$.

We repeat this preprocessing framework varying the time slot used to compute the crowd flows and the size of tiles in the tessellation, so to create different datasets. 
Specifically, we vary the time aggregation value in the set $\{15, 30, 45, 60\}$ minutes, and the tile size in the set $\{750, 1000, 1500\}$ meters for New York and Washington and $\{ 7500, 10000, 15000\}$ for Beijing.
Table \ref{tab:combination_time_intervals_tile_sizes} describes some statistics of the datasets and the map size after the preprocessing steps.

\begin{table}[htb!]
\centering
\resizebox{\columnwidth}{!}{%
\begin{tabular}{l|l|l|l|}
\cline{2-4}
                                                                                                       & \textbf{Bike NYC}                                                                          & \textbf{Taxi BJ}                                                                       & \textbf{Bike DC}                                                                              \\ \hline
\multicolumn{1}{|l|}{\textbf{Data Type}}                                                               & Bike Rent                                                                                  & GPS                                                                                              & Bike Rent                                                                                     \\ \hline
\multicolumn{1}{|l|}{\textbf{Location}}                                                                & New York City                                                                              & Beijing, China                                                                                   & Washington D.C.                                                                               \\ \hline
\multicolumn{1}{|l|}{\textbf{Timespan}}                                                                & 04--10 2014                                                                     & 02-2008                                                                               & 01-2018 -- 01-2020                                                                            \\ \hline
\multicolumn{1}{|l|}{\textbf{\begin{tabular}[c]{@{}l@{}}Spatial Agg.\end{tabular}}} & \begin{tabular}[c]{@{}l@{}}750 m (10 x 15)\\ 1000 m (7 x 11)\\ 1500 m (5 x 8)\end{tabular} & \begin{tabular}[c]{@{}l@{}}7500 m (32 x 32)\\ 10000 m (24 x 26)\\ 15000 m (16 x 16)\end{tabular} & \begin{tabular}[c]{@{}l@{}}750 m (32 x 40)\\ 1000 m (23 x 30)\\ 1500 m (14 x 18)\end{tabular} \\ \hline
\multicolumn{1}{|l|}{\textbf{Sampling}}                                                                & -                                                                                          & 177 sec.                                                                                         & -                                                                                             \\ \hline
\multicolumn{1}{|l|}{\textbf{\# Subjects}}                                                  & 421                                                                                        & 10,357                                                                                           & 557                                                                                           \\ \hline
\end{tabular} }
\caption{ \small Different information for the datasets used.}
\label{tab:combination_time_intervals_tile_sizes}
\end{table}

\subsection{Evaluation Metrics}
\label{sec:evaluation_metrics}
The performance of crowd flow predictors is evaluated as the similarity between the predicted heatmap of crowd flows and the real one. 
In our experiments, we adopt Root Mean Squared Error (RMSE), the most used metric to evaluate crowd flow prediction \cite{luca2021survey}:
$$\text{RMSE} = \sqrt{\frac{1}{n} \sum_{i=1}^{n} (y_i - \hat{y}_i)^{2}}$$
where $n$ is the number of predictions, $\hat{y}_i$ indicates the predicted value and $y_i$ the actual value.
We exploit the RMSE to evaluate the model performance on both the crowd flow prediction problem and on the flow prediction problem.

We evaluate the goodness of the predictions for the flow prediction problem using the Common Part of Commuters (CPC) \cite{cpc, luca2021survey, simini2021deep, LENORMAND2016158}:
\begin{equation}\label{eq:CPC}
CPC(\hat{T}, T) = \frac{2\sum_{i,j}min(\hat{T}_{ij}, T_{ij})}{\sum_{i,j}\hat{T_{ij}} + \sum_{ij}T_{ij}}
\end{equation}
where $\hat{T}_{ij}$ is the flow from region $i$ to region $j$ predicted by the model and $T_{ij}$ is the actual flow from region $i$ to region $j$.
CPC ranges between 0 and 1: if two adjacency matrices do not have any flows in common, CPC value is 0. 
CPC is 1 if the sets of flows are identical.

\subsection{Baselines}
We compare CrowdNet with the following baselines:

\begin{itemize}
    \item \textbf{Naïf approach}: the predicted crowd flows are the average of the crowd flows in the previous $n$ time slots;
    
    \item \textbf{Auto-Regressive Integrated Moving Average (ARIMA)}: a statistical model for understanding and forecast future values in a time series;
    
    \item \textbf{Vector Auto-Regressive (VAR)}: a variation of ARIMA that exploits multiple time series to capture the pairwise relationships among all flows;
    
    \item \textbf{ST-ResNet} \cite{Zhang2016}: a deep neural network prediction model for spatio-temporal data, which shows state-of-the-art results on crowd flows prediction.
    
    \item \textbf{DMVSTNet} \cite{YaoDMVSTNet}: framework able to model temporal view, spatial view, and semantic view, it models correlations among regions sharing similar temporal patterns. 
    
    \item \textbf{ACMF} \cite{LiuACMF} is a model able to infer the evolution of the crowd flow by learning dynamic representations of temporally-varying data exploiting an attention mechanism.
    
\end{itemize}

Table \ref{tab:comparison_baselines} shows the results of our model compared with these baselines on BikeNYC, BikeDC and TaxiBJ datasets with tiles of size 1000 meters and time intervals of 60 minutes. Using such tile size and time intervals, our model outperforms all the baselines taken into account with an RMSE score of 8.53 in the Bike NYC dataset, 14.91 in Taxi BJ dataset and 1.51 in Bike DC dataset. Similar results are obtained using deep learning models. On the other hand, statistic based methods have worse performance.

\begin{table}[]
\begin{tabular}{l|l l l}
\hline
Model         & Bike NYC & Taxi BJ  & Bike DC \\ \hline
Naïf approach & 19.87 & 46.12 & 2.64   \\
ARIMA         & 12.65 & 25.98 & 2.14  \\
VAR           & 12.50  & 25.64 & 1.88 \\
\hline
ST-ResNet     & 9.38  & 19.33 & 1.64  \\
DMVSTNet      & 9.16  & 18.89 & 1.60 \\
ACMF          & 8.89  & 18.28 & 1.55 \\
\bf CrowdNet        & \bf 8.53  & \bf 14.91 & \bf 1.51  
\end{tabular}
\caption{\small Comparisons of CrowdNet with baselines on BikeNYC, TaxiBJ and BikeDC.}
\label{tab:comparison_baselines}
\end{table}

\subsection{Experimental settings}
\label{sec:experimentalSettings}

We split each dataset into a development set and a test set. 
The development set includes a training set and a validation set.
80\% of the development set is considered as training set, and the remaining 20\% composes the validation set.
The test set contains the trips of the last ten days of the dataset. 

For ARIMA, we adopt the following hyperparameters: $p=12$ (where $p$ is the order of the autoregressive model), $d=0$ (where $d$ is the order of differentiation) and $q=24$ (the size of the moving average window).
For VAR, we use the following hyperparameter values: $p=8$, $d=0$ and $q=24$.
For STResNet, we use the same hyperparameter values as the original paper by Zhang et al. \cite{Zhang2016}, for the sake of clarity, they are reported in Table \ref{tab:hypterparameters}.

For CrowdNet, we perform a fine tuning of the hyperparameters using a grid search.
We select the hyperparameter values corresponding to the best performance obtained on the validation set (Table \ref{tab:hypterparameters}).


\begin{table}[htb!]
    \centering
    \begin{tabular}{|c|c|c|}
    \hline
            & CrowdNet & ST-ResNet \\ \hline
        Number of epochs & 150 & 100 \\ \hline
        Batch size & 16 & 32 \\ \hline
        Learning rate & 1e-4 & 2e-4 \\ \hline
        Optimiser & RMSprop & Adam \\ \hline
        Previous time intervals & 12 & 3+4+4
        \\ \hline
    \end{tabular}
    \caption{\small Hyperparameters for CrowdNet and ST-ResNet.}
    \label{tab:hypterparameters}
\end{table}


We build  CrowdNet and the baselines using PyTorch version 1.8.0 and we perform the experiments using a machine equipped with a Nvidia Quadro RTX 6000 as GPU (with 24 GB of GPU memory). 

We train the models adopting a validation-based early-stopping on each training dataset extracted from the flow datasets of Table \ref{tab:combination_time_intervals_tile_sizes} and on irregular tessellation for all the defined time intervals ($15min$, $30min$, $45min$ and $60min)$.

\section{Results}
\label{sec:results}

In this section, we compare the results of CrowdNet with those of ST-ResNet for the crowd flow prediction problem.
Moreover, the following Tables contain results related to the BikeNYC datasets. The same Tables and results for the datasets of BikeDC and TaxiBJ can be found in the Appendix. We report only the BikeNYC results as the behaviour of CrowdNet on the other datasets does not change.

\begin{figure}
    \centering
    \includegraphics[width=\columnwidth]{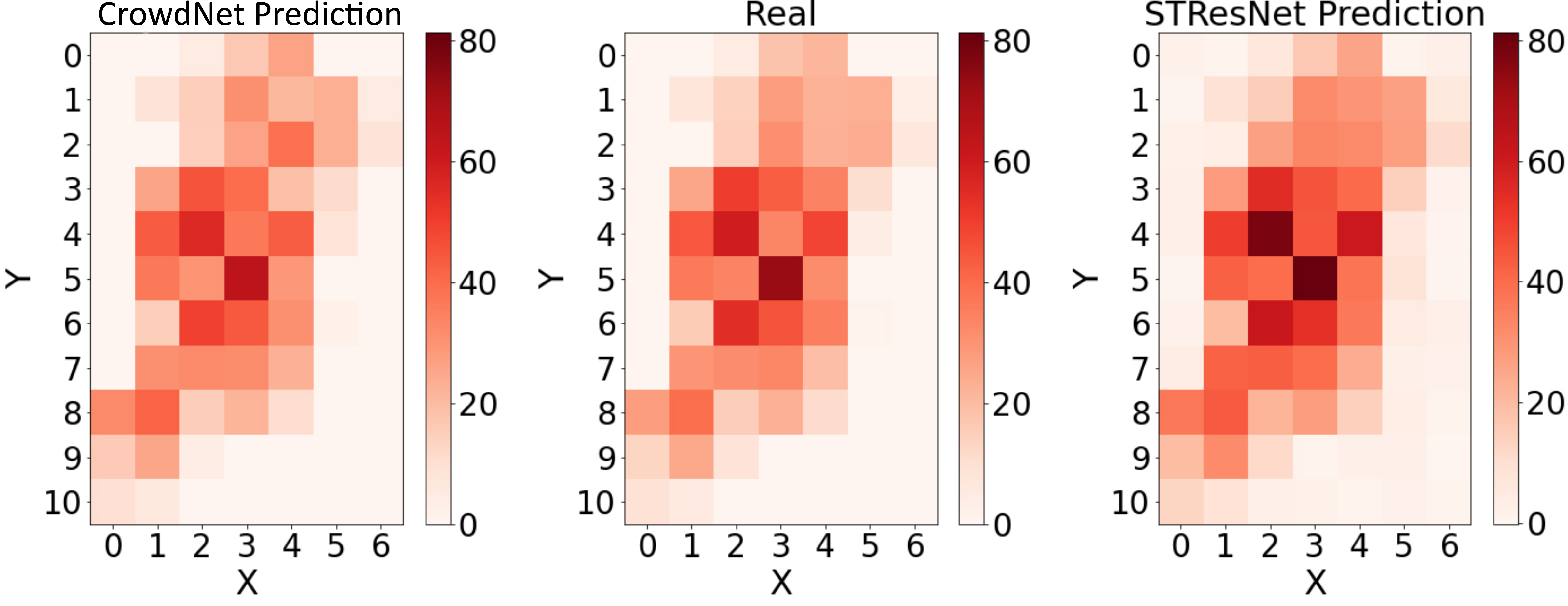}
    \caption{\small Comparison of mean real crowd inflows (center) with those predicted by CrowdNet (left) and STResNet (right).}
    \label{fig:Heatmap_Inflow_1000m_60min}
\end{figure}

Figure \ref{fig:Heatmap_Inflow_1000m_60min} visually compares the mean real crowd inflows with the mean crowd inflows predicted by STResNet and CrowdNet, with tessellations of $1000m$ and time interval of $60min$. 
It is evident how the predicted crowd flows are strikingly similar to the real ones. 
For example, the predictions well reproduce a notable pattern in Manhattan: the presence of areas with large crowd flows in the middle of the island, and the concentration of areas with small crowd flows in its borders.
In general, CrowdNet slightly underestimates large crowd flows. 

\begin{figure}[htb!]
    \centering
\begin{subfigure}{\columnwidth}
    \centering
    \includegraphics[width=\linewidth]{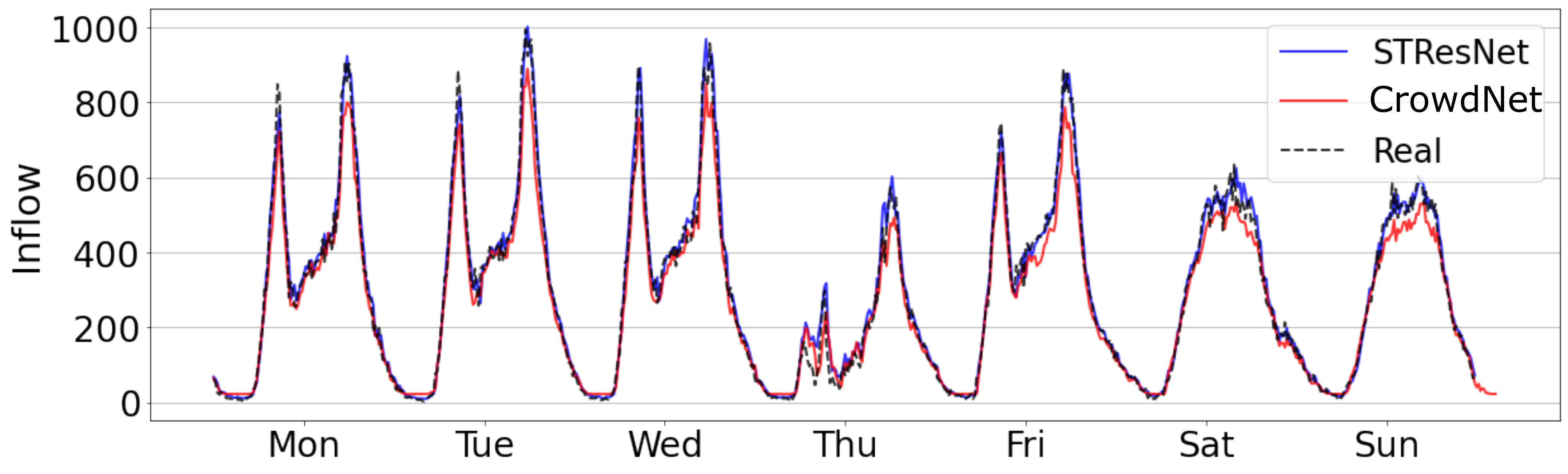}
    \caption{\small Crowd inflows: $1000m$, $15min$.}
    \label{fig:Timeseries_1000m_15min}
\end{subfigure}

\begin{subfigure}{\columnwidth}
    \centering
    \includegraphics[width=\linewidth]{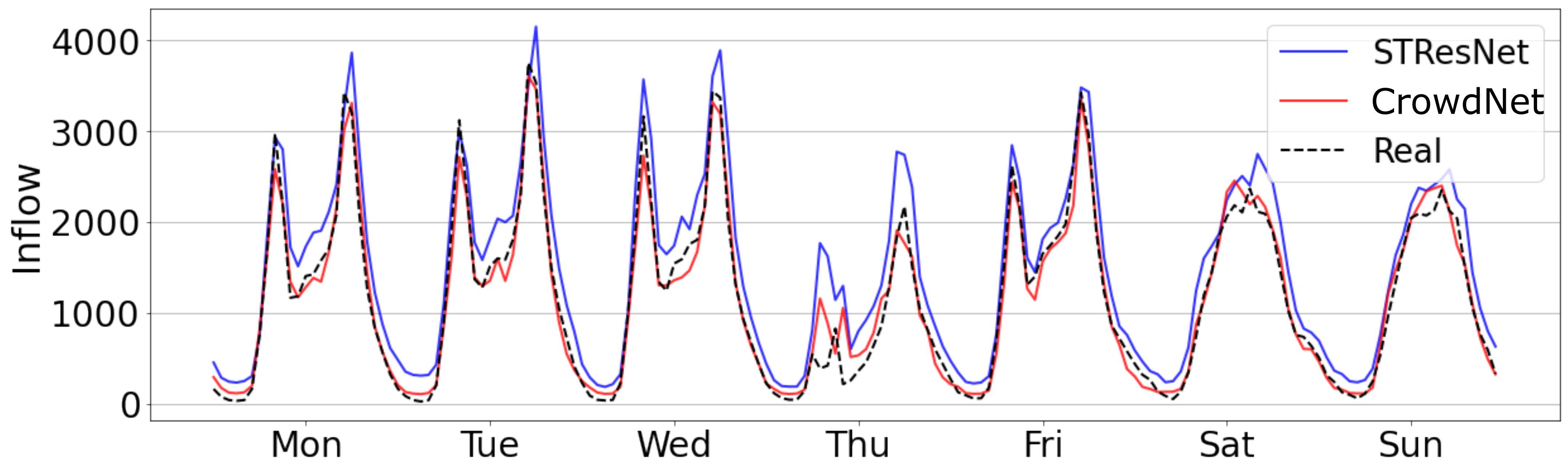}
    \caption{\small Crowd inflows: $1000m$, $60min$.}
    \label{fig:Timeseries_1000m_60min}
\end{subfigure}
    \caption{\small Comparison of the time series describing the total real crowd inflow in Manhattan with those resulting from the predictions of STResNet and CrowdNet.}
    \label{fig:Timeseries_1000m_15min_60min}
\end{figure}

Figure \ref{fig:Timeseries_1000m_15min} compares the sum of the total real crowd inflows during one week (from the 22nd to the 28th of Semptember), for tiles of $1000m$ and time intervals of $15min$, with those predicted by STResNet and CrowdNet. 
In general, all models underestimate large crowd flows and overestimate for low ones. 

Figure \ref{fig:Timeseries_1000m_60min} makes the comparison for a larger time interval ($60min$). 
In this case, CrowdNet's predictions are closer to the real values than STResNet's predictions, therefore we can say that CrowdNet performs better in case of high time intervals.
We find that, when external events occur such as the thunderstorm occurring on Thursday in Figure \ref{fig:Timeseries_1000m_60min}, the performance of the models is worse.

Table \ref{tab:results_CrowdNet_STResNet} summarises the results obtained in the test set by both models, reporting the RMSE for all the possible combination of tile sizes and time intervals. 
Note how CrowdNet performs better for larger time intervals, while for smaller values the RMSE of the two models is comparable.

\begin{table*}[htb!]
\resizebox{\linewidth}{!}{%
\begin{tabular}{cl|cccccccccccc|}
\cline{3-14}
\multicolumn{1}{l}{} & & \multicolumn{12}{c|}{Tile sizes}  \\ \cline{3-14} 
\multicolumn{1}{l}{} & & \multicolumn{4}{c|}{750m} & \multicolumn{4}{c|}{1000m} & \multicolumn{4}{c|}{1500m} \\ 
\cline{3-14}
\multicolumn{1}{l}{}            &       & \multicolumn{1}{c|}{CrowdNet} & \multicolumn{1}{c|}{STResNet} & \multicolumn{1}{l|}{DMVSTNet} & \multicolumn{1}{l|}{ACMF} & \multicolumn{1}{c|}{CrowdNet} & \multicolumn{1}{c|}{STResNet} & \multicolumn{1}{l|}{DMVSTNet} & \multicolumn{1}{l|}{ACMF} & \multicolumn{1}{c|}{CrowdNet} & \multicolumn{1}{c|}{STResNet} & \multicolumn{1}{l|}{DMVSTNet} & \multicolumn{1}{l|}{ACMF} \\ \hline

\multicolumn{1}{|c|}{} & 15min & \multicolumn{1}{c|}{1.71}     & \multicolumn{1}{c|}{1.69}     & \multicolumn{1}{c|}{1.65}     & \multicolumn{1}{c|}{\textbf{1.63}} & \multicolumn{1}{c|}{2.76}     & \multicolumn{1}{c|}{2.35}     & \multicolumn{1}{c|}{2.29}     & \multicolumn{1}{c|}{\textbf{2.26}} & \multicolumn{1}{c|}{3.73}     & \multicolumn{1}{c|}{3.35}     & \multicolumn{1}{c|}{3.27}     & \textbf{3.23}                      \\ \cline{2-14} 
\multicolumn{1}{|c|}{Time}      & 30min & \multicolumn{1}{c|}{3.69}     & \multicolumn{1}{c|}{2.65}     & \multicolumn{1}{c|}{2.59}     & \multicolumn{1}{c|}{\textbf{2.55}} & \multicolumn{1}{c|}{5.23}     & \multicolumn{1}{c|}{4.85}     & \multicolumn{1}{c|}{4.73}     & \multicolumn{1}{c|}{\textbf{4.67}} & \multicolumn{1}{c|}{5.93}     & \multicolumn{1}{c|}{5.64}     & \multicolumn{1}{c|}{5.51}     & \textbf{5.44}                      \\ \cline{2-14} 
\multicolumn{1}{|c|}{intervals} & 45min & \multicolumn{1}{c|}{4.34}     & \multicolumn{1}{c|}{3.67}     & \multicolumn{1}{c|}{3.58}     & \multicolumn{1}{c|}{\textbf{3.54}} & \multicolumn{1}{c|}{6.68}     & \multicolumn{1}{c|}{5.63}     & \multicolumn{1}{c|}{5.50}     & \multicolumn{1}{c|}{\textbf{5.43}} & \multicolumn{1}{c|}{11.3}     & \multicolumn{1}{c|}{10.91}    & \multicolumn{1}{c|}{10.66}    & \textbf{10.53}                     \\ \cline{2-14} 
\multicolumn{1}{|c|}{}          & 60min & \multicolumn{1}{c|}{\textbf{5.18}}     & \multicolumn{1}{c|}{5.44}     & \multicolumn{1}{c|}{5.31}     & \multicolumn{1}{c|}{5.25} & \multicolumn{1}{c|}{\textbf{8.53}}     & \multicolumn{1}{c|}{9.38}     & \multicolumn{1}{c|}{9.16}     & \multicolumn{1}{c|}{8.89} & \multicolumn{1}{c|}{\textbf{11.1}}     & \multicolumn{1}{c|}{11.66}    & \multicolumn{1}{c|}{11.39}    & 11.25                     \\ \hline
\end{tabular}
}
\caption{\small Performance of ST-ResNet, DMVSTNet, ACMP and CrowdNet models for crowd flow prediction problem on the BikeNYC dataset.}
\label{tab:results_CrowdNet_STResNet}
\end{table*}

One of the advantages of using CrowdNet is the possibility to use irregular tessellations.
STResNet, DMBSTNet and ACMP cannot be used with irregular tessellations because they take as input an image-like matrix; irregular tessellations cannot be represented easily in this way. 
For instance, considering an administrative tessellation, a region can have different neighbours, and it is difficult to represent this kind of relationship using an image-like relationship.

We illustrate the performance of CrowdNet on an irregular tessellation defined by an administrative tessellation defined by 29 neighbourhoods in Manhattan. Data of the administrative tiles are taken from an official tool of the municipality of New York City \footnote{\url{lhttps://popfactfinder.planning.nyc.gov/}}.

\begin{figure}[!]
\centering
    \centering
    \includegraphics[width=\linewidth]{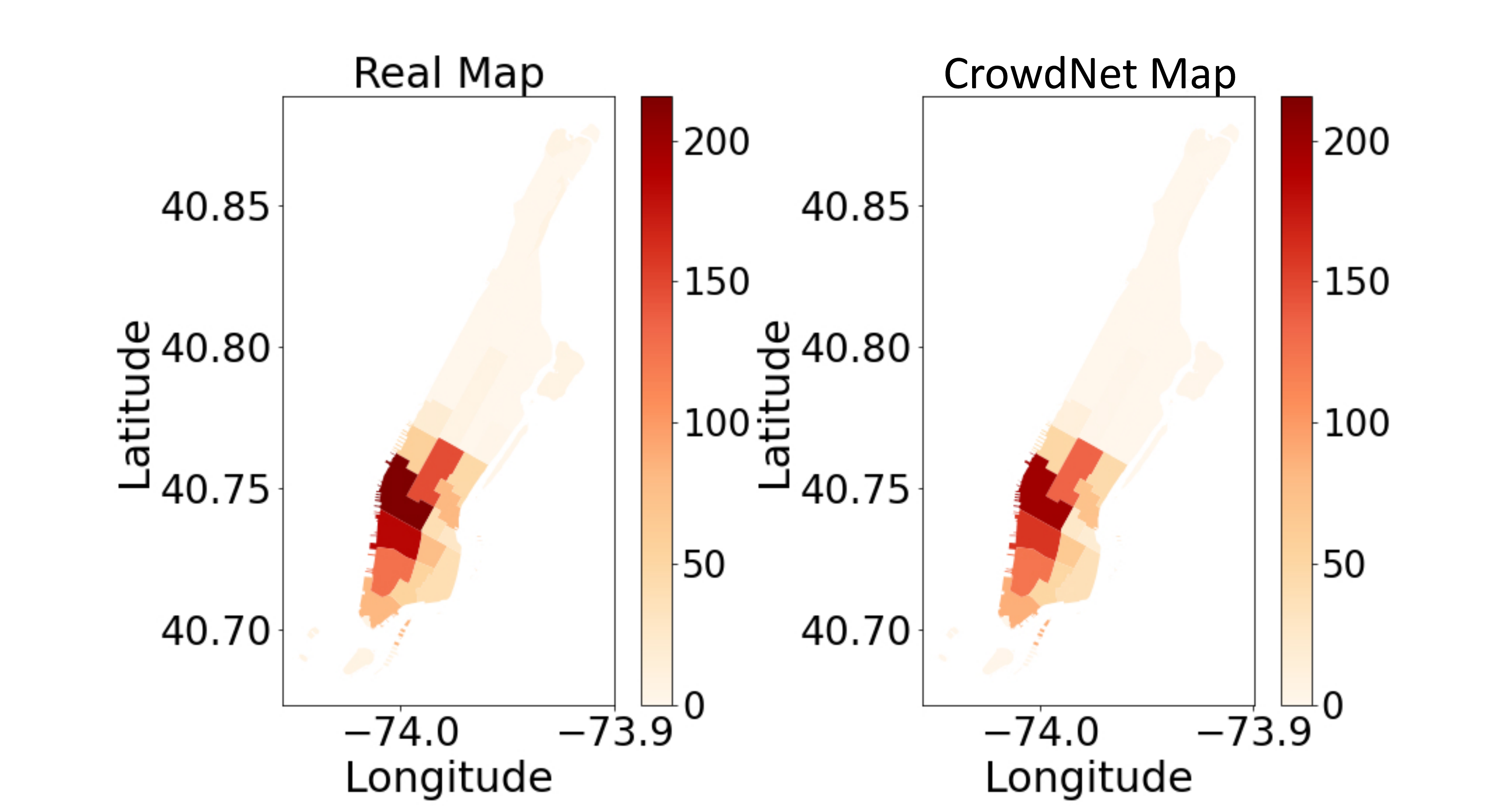}
    \caption{\small Comparison of mean real crowd outflows (left) with those predicted by CrowdNet (right). The crowd flows are represented as heatmaps, in which the colour of each cell is proportional to the crowd flow of the corresponding tile.} \label{fig:Heatmap_AdministrativeArean}
\end{figure}

Figure \ref{fig:Heatmap_AdministrativeArean} compares, the real crowd inflows and outflows with the crowd inflows and outflows predicted by CrowdNet.
As in the case of the squared tessellation, CrowdNet's predictions are strikingly similar to the real one. 
Note how, also in this case, larger crowd inflows and outflows are concentrated in the middle of the island, while smaller ones concentrate in the southern part of the island.

\paragraph{Flow prediction}
While crowd flow prediction aims to forecast the aggregated flows in each tile, flow prediction aims at predicting the flow between each pair of tiles, thus corresponding to the prediction of the entire origin-destination matrix.

\begin{table}[htb!]
\begin{tabular}{cl|c|c|c|}
\cline{3-5}
\multicolumn{1}{l}{}                 &       & \multicolumn{3}{c|}{Tile sizes} \\ \cline{3-5} 
\multicolumn{1}{l}{}                 &       & 750m      & 1000m    & 1500m    \\ \hline
\multicolumn{1}{|c|}{} & 15min & 0.248     & 0.409    & 0.622    \\ \cline{2-5} 
\multicolumn{1}{|c|}{Time}               & 30min & 0.357     & 0.654    & 1.014    \\ \cline{2-5} 
\multicolumn{1}{|c|}{intervals}               & 45min & 0.460     & 0.880    & 1.396    \\ \cline{2-5} 
\multicolumn{1}{|c|}{}               & 60min & 0.538     & 1.049    & 1.815    \\ \hline
\end{tabular}
\caption{\small Performance of CrowdNet model for flow prediction on the BikeNYC test set in terms of RMSE.}
\label{tab:results_CrowdNet_RMSE}
\end{table}

\begin{table}[]
\begin{tabular}{cl|c|c|c|}
\cline{3-5}
\multicolumn{1}{l}{}                 &       & \multicolumn{3}{c|}{Tile sizes} \\ \cline{3-5} 
\multicolumn{1}{l}{}                 &       & 750m      & 1000m    & 1500m    \\ \hline
\multicolumn{1}{|c|}{} & 15min & 0.106     & 0.218    & 0.414    \\ \cline{2-5} 
\multicolumn{1}{|c|}{Time}               & 30min & 0.193     & 0.368    & 0.559    \\ \cline{2-5} 
\multicolumn{1}{|c|}{intervals}               & 45min & 0.274     & 0.460    & 0.631    \\ \cline{2-5} 
\multicolumn{1}{|c|}{}               & 60min & 0.321     & 0.496    & 0.637    \\ \hline
\end{tabular}
\caption{\small Performance of CrowdNet on the BikeNYC test set in terms of CPC.}
\label{tab:results_CrowdNet_CPC}
\end{table}

\begin{figure}
    \centering
    \includegraphics[width=\columnwidth]{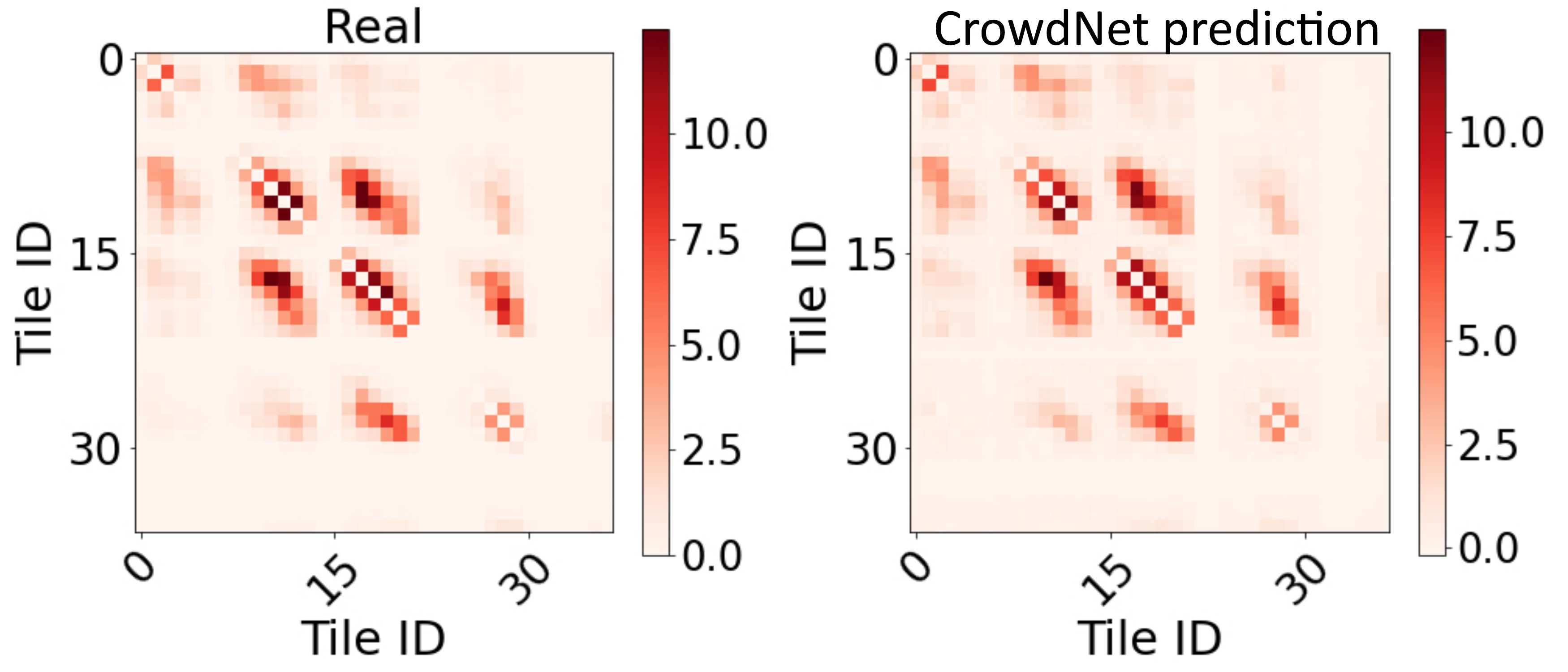}
    \caption{\small Representation of real adjacency matrices and the ones predicted by CrowdNet for BikeNYC dataset with tile size: $1500m$. Time interval: $60min$.}
    \label{fig:HeatmapAdjacencyMatrix_1500_60min}
\end{figure}

Figure \ref{fig:HeatmapAdjacencyMatrix_1500_60min} compares the real flows with those predicted by CrowdNet: the two predictions are almost identical, meaning that the flows are correctly predicted by the model.

\begin{figure}
\centering
\begin{subfigure}{.45\columnwidth}
    \centering
    \includegraphics[width=\linewidth]{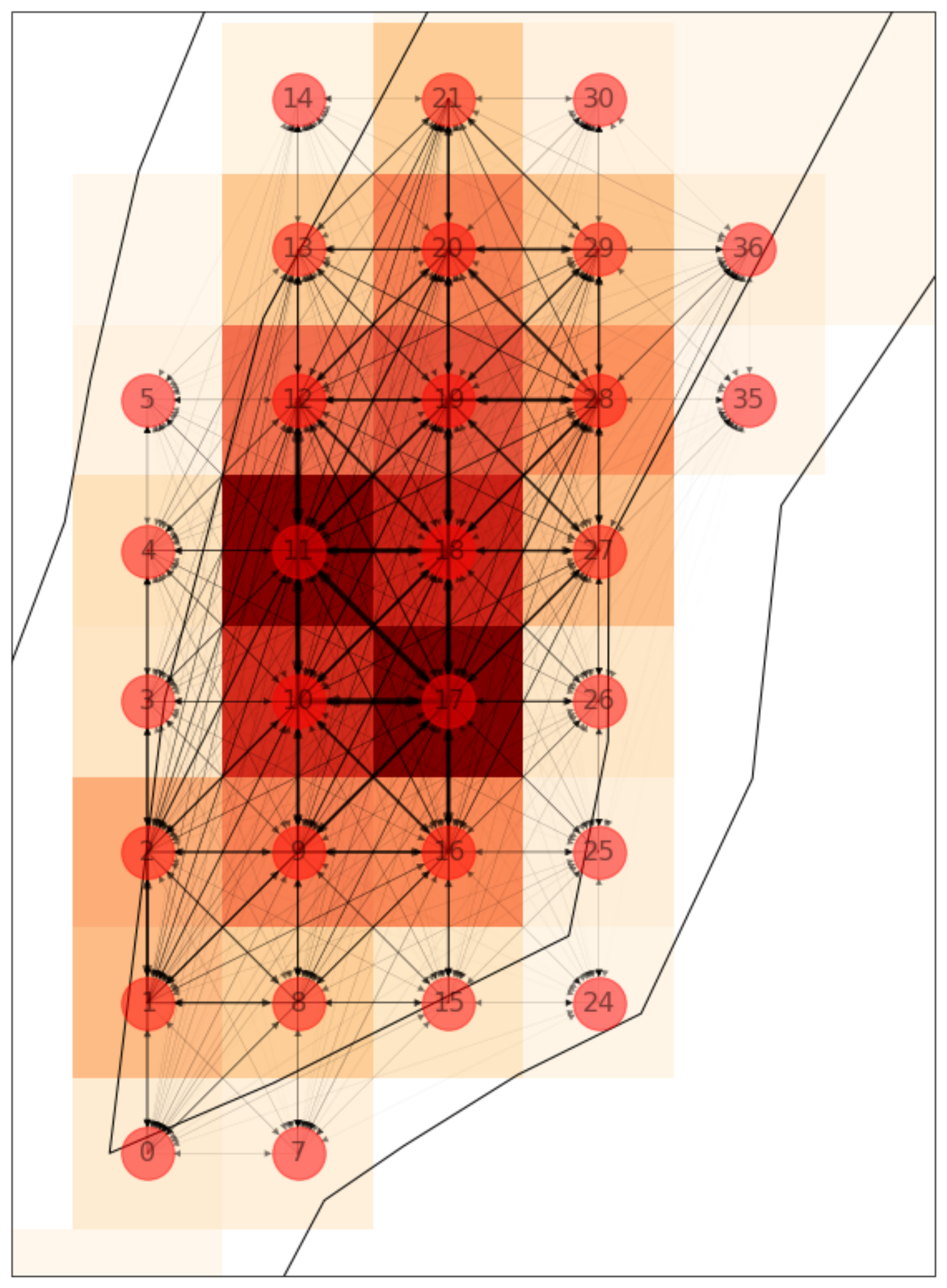}
    \caption{\small Network of flows between all tiles.}
    \label{fig:Manhattan_inflow_and_outflows}
\end{subfigure}
\begin{subfigure}{.45\columnwidth}
    \centering
    \includegraphics[width=\linewidth]{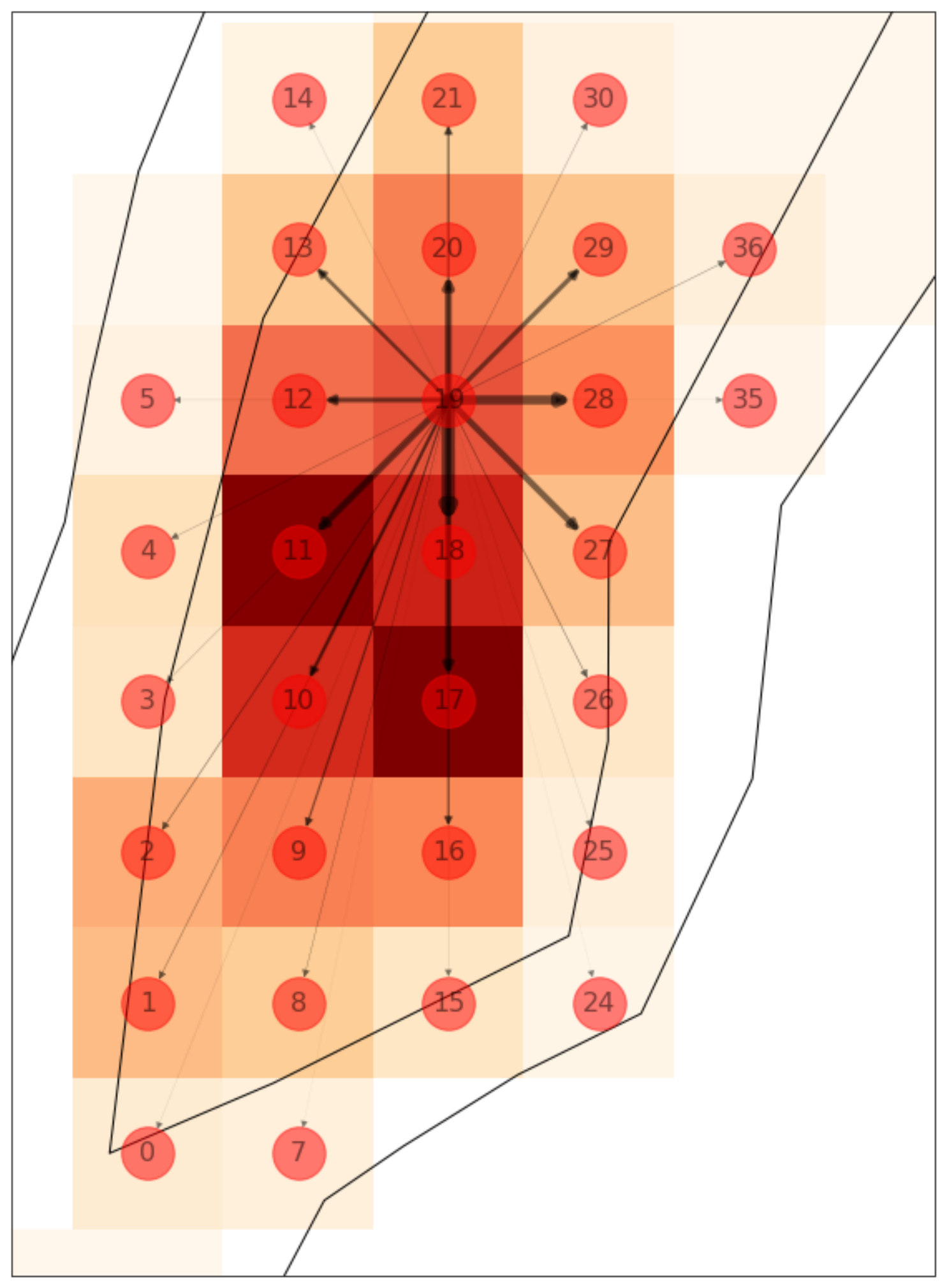}
    \caption{\small Flows outgoing from a single tile (node 17).}
    \label{fig:Manhattan_flows_node17}
\end{subfigure}

    \caption{\small Crowd flows prediction enriched with flow network information. 
    Each edge in the network represents a flow between a pair of tiles, with the thickness of edges proportional to the flow value. 
    The heatmap in the background represents the crowd flow prediction.}
    \label{fig:Manhattan_inflow_and_outflows_flow}
\end{figure}

Providing detailed information about the crowd flow predictions is essential to acquire knowledge that can be useful to possible users, such as policymakers and urban planners.
In this direction, CrowdNet enriches the predicted crowd flows with useful information, such as the origin and the destination of each flow.
As an example, Figure \ref{fig:Manhattan_inflow_and_outflows} illustrates the predicted crowd flows in companion with the flows between the tiles in Manhattan: each node's size is proportional to its crowd inflow, while edge thickness represents the magnitude of the single flows between pairs of tiles. 
Figure \ref{fig:Manhattan_flows_node17} shows the same crowd flow prediction with a focus on the flows outgoing from node \textit{17}.
The ability of CrowdNet to solve flow prediction allows us to enrich significantly the crowd flow prediction output with information about the origin of a tile's inflow or outflow.

To investigate the robustness of CrowdNet, for the flow prediction problem, to variation in the spatial and temporal aggregation, we investigate how the performance of CrowdNet changes varying time intervals (fixing a tile size of $1000m$) or tile sizes (fixing time intervals to $60min$).\footnote{The results fixing others time intervals or tile sizes are similar.}

\begin{figure}[htb!]
    \centering
    \begin{subfigure}{0.45\columnwidth}
    \centering
    \includegraphics[width=\columnwidth]{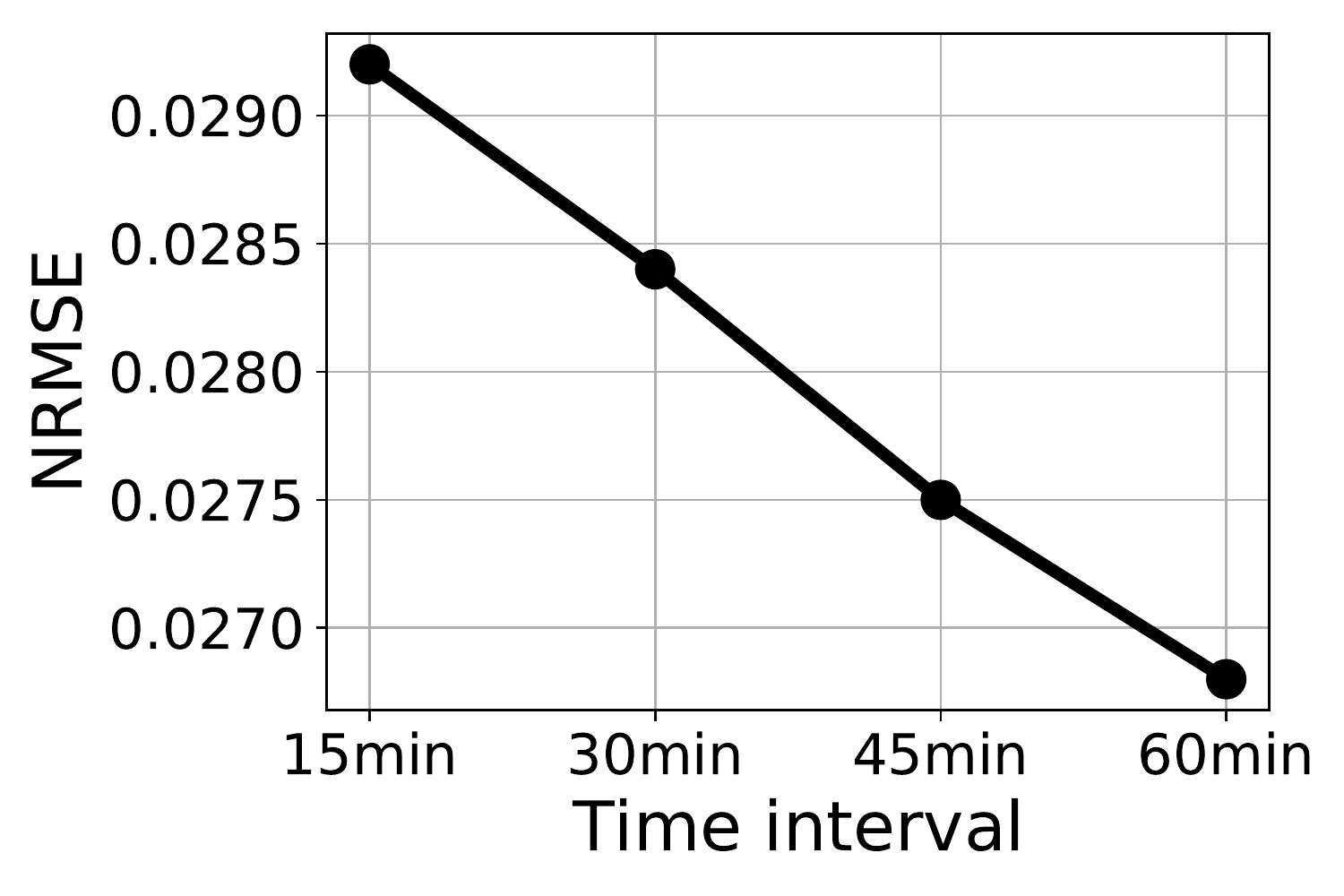}
    \caption{\small NRMSE for a fixed tile size of $1000m$}
    \label{fig:NRMSE_fixed_tile_size}
    \end{subfigure}
    \begin{subfigure}{0.45\columnwidth}
    \includegraphics[width=\columnwidth]{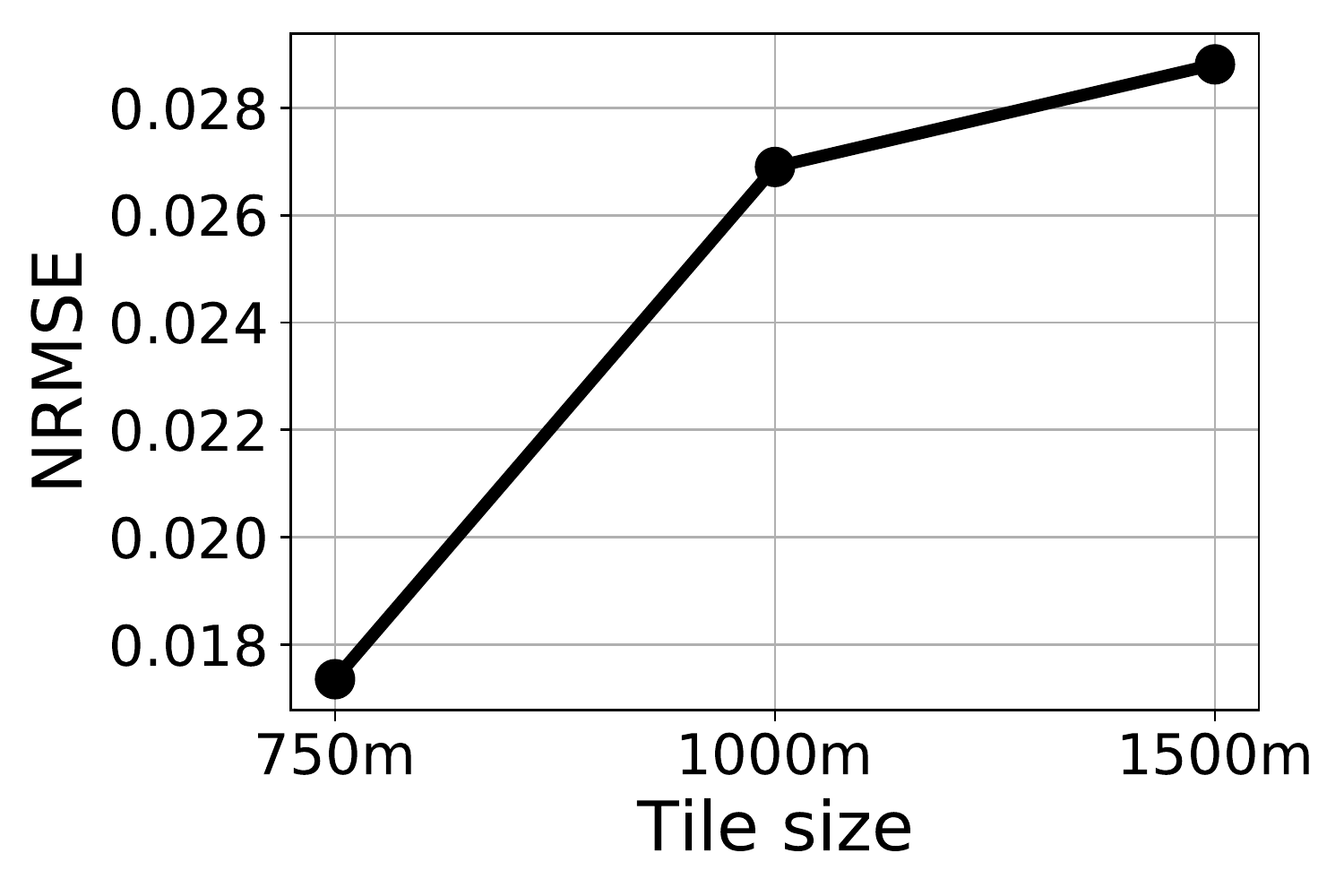}
    \caption{\small NRMSE for a fixed time interval of $60min$.}
    \label{fig:NRMSE_fixed_time_interval}
    \end{subfigure}
    \caption{\small NRMSE for flow prediction problem varying time intervals and tile sizes.}
    \label{fig:NRMSE}
\end{figure}

CrowdNet is robust with respect to the temporal aggregation. As shown by Table \ref{tab:results_CrowdNet_RMSE} the RMSE increases as the time interval increases. This is due to the fact that the flow magnitude increases as the time intervals becomes bigger.
Normalising the RMSE with respect to the maximum flow allows us to evaluate the model on different tile sizes and time intervals without considering the error's magnitude. 
Figures \ref{fig:NRMSE_fixed_tile_size} and \ref{fig:NRMSE_fixed_time_interval} show the Normalised RMSE of the model, defined as: 
$$
    NRMSE = \frac{RMSE}{f_{max}-f_{min}}
$$
where $f_{max}$ is the maximum value of the flow and $f_{min}$ is the minimum value.
We observe that, as the time intervals increase, CrowdNet's NRMSE decreases. This behaviour is shown in Figure \ref{fig:NRMSE_fixed_tile_size}.

Similarly, considering the spatial aggregation, the RMSE increases as the tile size increases. This is due to the fact that as the tile size becomes bigger its flow increases.
In the case of spatial aggregation, the NRMSE increases as the tile size increases. This behaviour is shown in figure \ref{fig:NRMSE_fixed_time_interval}.

\begin{figure}
    \centering
    \includegraphics[width=0.7\columnwidth]{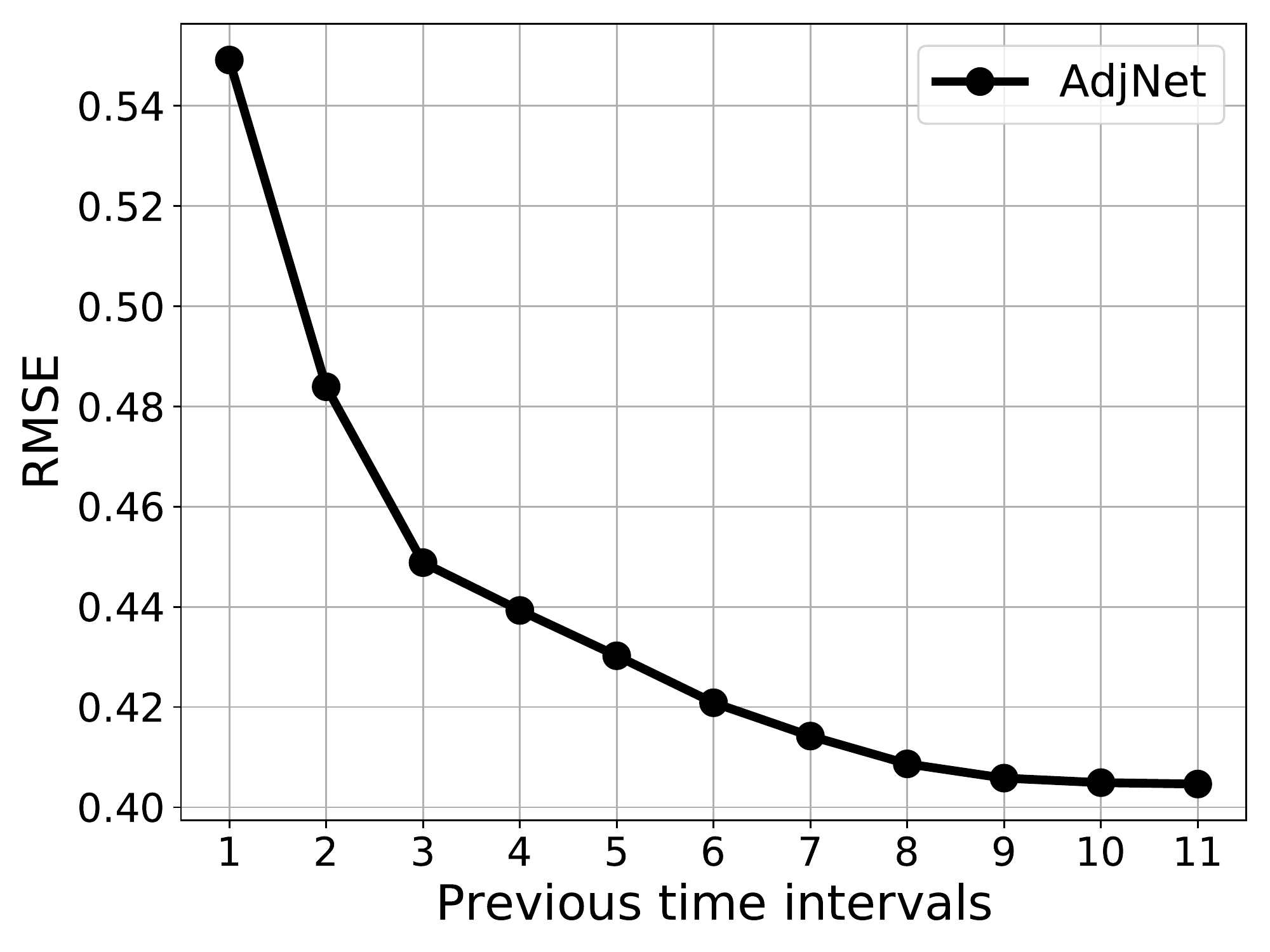}
    \caption{\small Performance of CrowdNet in terms of RMSE in the BikeNYC dataset for tile size $1000m$ and time intervals of $60min$.}
    \label{fig:temporal_importance}
\end{figure}

\paragraph{Temporal importance}
Figure \ref{fig:temporal_importance} shows how the performance of CrowdNet in the BikeNYC dataset changes based on the number of previous time intervals used to make the prediction. 
It is possible to observe that an higher number of time intervals used to make the prediction leads to better performance in terms of RMSE. The plot shows also a plateau around the 11th time interval.

\section{Discussion and Conclusions}
\label{sec:discussion_conclusions}

Our paper focused on crowd flow prediction, i.e., forecasting the number of people that leave (or enter) a region in a geographic area.
We proposed a novel solution to this task called CrowdNet, which represents crowd flows by means of a graph represented as an adjacency matrix, in which nodes represent regions in the geographic area and edges represent people moving among regions. 
The usage of a graph-based model provides many advantages for policymakers. For instance, our solution allows policymakers to use spatial-tessellation (e.g., make predictions at street level, block-level, and others). Such versatility may enable policymakers to use our model to solve various modern challenges. As an example, while previous models allow predicting flows only on squared grids, making predictions of flows at a street level is more helpful to take countermeasures to environmental problems (e.g., pollution). Similarly, for other social challenges like crime prevention, policymakers can exploit our model to predict at a block level, gathering better insights than those derived from squared tessellations.
Technically, CrowdNet's predictions are directed edges flows predictions on the graph representing crowds movements.
This allows to perform predictions also on non-squared tessellation, representing one of the novelties proposed within our approach.
CrowdNet's predictions may be then aggregated so to obtain a solution to crowd flow prediction.
Finally, another key contribution of our work consists in the characterisation of the behaviour of CrowdNet on different types of tessellation with different shapes and sizes and on different time intervals. 
This allowed us to find what are the combinations of tessellation and time interval which lead to the most accurate predictions.
The performance of CrowdNet is comparable to that of STResNet, confirming the quality of our contribution, with the difference that CrowdNet degrades as flows magnitude decrease. 
This is due to the fact that, contrarily to STResNet, CrowdNet does not predict directly crowd inflows and outflows. Instead, the prediction is derived from many single predictions performed on every flow. 
Our model, together with the proposed analysis, is intended to be a first step towards the adoption of more exhaustive prediction models. Our experiments may lead policymakers to the adoption of more fair, transparent, trustable solutions thanks to their enriched prediction and parameterisation.
As a future improvement, it would be interesting to observe if its performance increases considering also previous days and weeks and implementing a fusion mechanism to make predictions.


\begin{acks}
Luca Pappalardo has been partially supported by EU project SoBigData++ grant agreement 871042.
\end{acks}

\bibliographystyle{ACM-Reference-Format}
\bibliography{sample-base}



\clearpage

\appendix
\section{Results on other datasets}
\setcounter{table}{10}
\begin{table}[!b]
\begin{tabular}{ll|c|c|c|c|c|c|}
\cline{3-8} &       & \multicolumn{6}{c|}{Tile sizes} \\ \cline{3-8} & & \multicolumn{2}{c|}{750m} & \multicolumn{2}{c|}{1000m} & \multicolumn{2}{c|}{1500m} \\ \cline{3-8}  & & \multicolumn{1}{c|}{CrowdNet} & \multicolumn{1}{c|}{STResNet} & \multicolumn{1}{c|}{CrowdNet} & \multicolumn{1}{c|}{STResNet} & \multicolumn{1}{c|}{CrowdNet} & \multicolumn{1}{c|}{STResNet} \\ \hline
\multicolumn{1}{|c|}{} & 15min & 9.82 & 9.88 & 11.54 & 12.27 & 15.91 & 15.88 \\ \cline{2-8} 
\multicolumn{1}{|c|}{Time} & 30min & 10.33 & 11.75 & 13.13 & 13.66 & 16.48 & 17.02 \\ \cline{2-8} 
\multicolumn{1}{|c|}{intervals} & 45min & 10.97 & 10.86 & 13.44 & 15.71 & 16.69 & 19.22 \\ \cline{2-8} 
\multicolumn{1}{|c|}{} & 60min & 12.78 & 15.34 & 14.91 & 19.33 & 18.53 & 24.37 \\ \hline
\end{tabular}

\begin{tabular}{ll|c|c|c|c|c|c|}
\cline{3-8} &       & \multicolumn{6}{c|}{Tile sizes} \\ \cline{3-8} & & \multicolumn{2}{c|}{750m} & \multicolumn{2}{c|}{1000m} & \multicolumn{2}{c|}{1500m} \\ \cline{3-8}  & & \multicolumn{1}{c|}{CrowdNet} & \multicolumn{1}{c|}{STResNet} & \multicolumn{1}{c|}{CrowdNet} & \multicolumn{1}{c|}{STResNet} & \multicolumn{1}{c|}{CrowdNet} & \multicolumn{1}{c|}{STResNet} \\ \hline
\multicolumn{1}{|c|}{} & 15min & 0.86 & 0.78 & 1.03 & 1.01 & 1.36 & 1.45 \\ \cline{2-8} 
\multicolumn{1}{|c|}{Time} & 30min & 0.94 & 0.96 & 1.17 & 1.21 & 1.54 & 1.57 \\ \cline{2-8} 
\multicolumn{1}{|c|}{intervals} & 45min & 1.02 & 1.08 & 1.32 & 1.49 & 2.01 & 2.14 \\ \cline{2-8} 
\multicolumn{1}{|c|}{} & 60min & 1.20 & 1.36 & 1.51 & 1.64 & 2.46 & 2.89 \\ \hline
\end{tabular}
\caption{\small Performance of the ST-ResNet and CrowdNet models for crowd flow prediction problem on the TaxiBJ (upper) and BikeDC (lower) test sets in terms of RMSE, varying the tile size and time interval.}
\label{tab:results_CrowdNet_STResNet_BJ}
\end{table}

\setcounter{table}{6}

\begin{table}[H]
\begin{tabular}{cl|c|c|c|}
\cline{3-5}
\multicolumn{1}{l}{}                 &       & \multicolumn{3}{c|}{Tile sizes} \\ \cline{3-5} 
\multicolumn{1}{l}{}                 &       & 750m      & 1000m    & 1500m    \\ \hline
\multicolumn{1}{|c|}{} & 15min & 1.387     & 1.729    & 1.891    \\ \cline{2-5} 
\multicolumn{1}{|c|}{Time}               & 30min & 1.816     & 2.174    & 2.438    \\ \cline{2-5} 
\multicolumn{1}{|c|}{intervals}               & 45min & 1.948     & 2.642    & 3.003    \\ \cline{2-5} 
\multicolumn{1}{|c|}{}               & 60min & 2.131     & 3.428    & 4.193    \\ \hline
\end{tabular}
\caption{\small Performance of CrowdNet model for \emph{flow} prediction on the Taxi Beijing test set in terms of RMSE, varying the tile size and time interval.}
\label{tab:results_CrowdNet_RMSE_BJ}
\end{table}

\begin{table}[H]
\begin{tabular}{cl|c|c|c|}
\cline{3-5}
\multicolumn{1}{l}{}                 &       & \multicolumn{3}{c|}{Tile sizes} \\ \cline{3-5} 
\multicolumn{1}{l}{}                 &       & 750m      & 1000m    & 1500m    \\ \hline
\multicolumn{1}{|c|}{} & 15min & 0.214     & 0.319    & 0.414    \\ \cline{2-5} 
\multicolumn{1}{|c|}{Time}               & 30min & 0.273     & 0.381    & 0.425    \\ \cline{2-5} 
\multicolumn{1}{|c|}{intervals}               & 45min & 0.301     & 0.429    & 0.572    \\ \cline{2-5} 
\multicolumn{1}{|c|}{}               & 60min & 0.387     & 0.562    & 0.714    \\ \hline
\end{tabular}
\caption{\small Performance of CrowdNet on the Taxi Beijing test set in terms of CPC, varying the tile size and time interval.}
\label{tab:results_CrowdNet_CPC_BJ}
\end{table}

\begin{table}[H]
\begin{tabular}{cl|c|c|c|}
\cline{3-5}
\multicolumn{1}{l}{}                 &       & \multicolumn{3}{c|}{Tile sizes} \\ \cline{3-5} 
\multicolumn{1}{l}{}                 &       & 750m      & 1000m    & 1500m    \\ \hline
\multicolumn{1}{|c|}{} & 15min  & 0.017 & 0.021   & 0.023    \\ \cline{2-5} 
\multicolumn{1}{|c|}{Time}               & 30min & 0.019     & 0.024    & 0.028    \\ \cline{2-5} 
\multicolumn{1}{|c|}{intervals}               & 45min & 0.031     & 0.044    & 0.049    \\ \cline{2-5} 
\multicolumn{1}{|c|}{}               & 60min & 0.034     & 0.045    & 0.53    \\ \hline
\end{tabular}
\caption{\small Performance of CrowdNet model for \emph{flow} prediction on the Bike Washington D.C. test set in terms of RMSE, varying the tile size and time interval.}
\label{tab:results_CrowdNet_RMSE_DC}
\end{table}

\begin{table}[H]
\begin{tabular}{cl|c|c|c|}
\cline{3-5}
\multicolumn{1}{l}{}                 &       & \multicolumn{3}{c|}{Tile sizes} \\ \cline{3-5} 
\multicolumn{1}{l}{}                 &       & 750m      & 1000m    & 1500m    \\ \hline
\multicolumn{1}{|c|}{} & 15min & 0.231     & 0.358    & 0.416    \\ \cline{2-5} 
\multicolumn{1}{|c|}{Time}               & 30min & 0.272     & 0.398    & 0.453    \\ \cline{2-5} 
\multicolumn{1}{|c|}{intervals}               & 45min & 0.429     & 0.596    & 0.721    \\ \cline{2-5} 
\multicolumn{1}{|c|}{}               & 60min & 0.512     & 0.657    & 0.801    \\ \hline
\end{tabular}
\caption{\small Performance of CrowdNet on the Bike Washington D.C. test set in terms of CPC, varying the tile size and time interval.}
\label{tab:results_CrowdNet_CPC_DC}
\end{table}

{









}
\end{document}